\documentclass[10pt]{article}

\usepackage[utf8]{inputenc}
\usepackage{amsmath}
\usepackage{amssymb}

\usepackage{geometry}
\usepackage{amsthm}       
\usepackage{dsfont}       
\usepackage{graphicx}     
\usepackage{stmaryrd}     
\usepackage{color}
\usepackage{subeqnarray}
\usepackage{subfigure}

\usepackage{algpseudocode}
\usepackage{algorithm2e}
\usepackage{algpascal}
\usepackage{authblk}

\usepackage[pdftex,linkcolor=test,citecolor=vertsombre,colorlinks=true,pagebackref,bookmarks=true, plainpages=true,urlcolor=freeblue]{hyperref}

\definecolor{test}{rgb}{0.64,0.16,0.16}
\definecolor{vertsombre}{rgb}{0.00,0.57,0.1}
\definecolor{freeblue}{rgb}{0.25,0.41,0.88}

\title{Constraint Reductions} 

\author[1]{Olivier Bailleux\thanks{ \texttt{olivier.bailleux@u-bourgogne.fr}}}
\author[,2]{Yacine Boufkhad\thanks{ \texttt{yacine.boufkhad@irif.fr}}}

\affil[1]{LIB Université de Bourgourgne}
\affil[2]{IRIF CNRS-Université Paris Diderot
}

\date{}

\begin{document}

\setlength{\parskip}{0.2cm}

\maketitle

\begin{abstract}
In this commentary on the \cite{BB03} paper, after recalling its context, we outline a classification of Constraints with respect to their deductive power regarding General Arc Consistency (GAC). 
\end{abstract}

\section{Introduction}

It is not usual to be asked to write something on a subject on which you have worked fifteen years before and on which you have remained far from subsequent developments. Yet this is the challenge we humbly accepted to undertake in this paper with the limitation that our expertise in this research area have diminished regarding the developments made by many others during the ten previous years. 

We begin with the easy part consisting in recalling the context of the paper \cite{BB03}.

In the sequel the word \emph{constraint} will be used in two meanings: a type of constraints, such as propositional clause, Boolean cardinality constraint, ALLDIFF..., and a constraint instance, i.e., the formal representation of a relation on several variables each having a domain. We will use the word \emph{Constraint} in the first case and \emph{constraint} in the second case. For example, CNF is a Constraint that denotes the set of propositional clauses, whereas $(a \vee \neg b \vee c)$ is a CNF constraint.

The main result in \cite{BB03} was twofold: the description of a general method of solving constraints satisfaction problems through translation and the application of this method to the translation of Boolean cardinality Constraint into CNF. 

The general method can be quickly described as follows: We have a constraint network based on one or more types of so called \textit{source Constraints}, and we want to solve this network using a solver dedicated to a type of so called \textit{target Constraint}. Then we need to translate each source constraint to an equivalent set  of target constraints. It is therefore desirable that the problem can be solved at least as effectively with the target solver as it would have been with an up to date source solver. The three most obvious criteria for achieving this objective are:
\begin{enumerate}
\item The translation process must be fast compared to the assumed resolution time of the source solver, namely, the translation algorithms must have a polynomial time complexity.
\item The target constraint network must not be of a prohibitive size.
\item All deductions that could be made by an up to date solver from source constraints must be feasible by the target solver.
\end{enumerate}

Let us call for the moment a translation that fulfills the three above criteria a good translation. Regarding the third criterion, almost all modern constraint solvers at least enforce domain consistency, also called generalized arc consistency or GAC. We therefore considered, as \cite{Gent2002ArcCI} before us, and as many authors thereafter, that it was desirable that enforcing domain consistency on target constraints produces at least the same deductions as enforcing domain consistency on source constraints.

We applied this method to the case of cardinality constraints as a source Constraint and clauses in CNF form as a target Constraint.
Most of SAT solvers use unit propagation as a deduction method and therefore we required for the third criterion that the translation restores the generalized arc consistency through unit propagation. At the time, it was known that a direct translation of cardinality constraints to CNF that preserves the power of deduction without adding any new variable requires an exponential number of clauses. Then a linear time translation was proposed in \cite{WARNERS199863}, producing a polynomial number of clauses, but at the cost of a loss of deduction power since the unit propagation does not restore arc consistency.

We proposed then in \cite{BB03} a good encoding (thus fulfilling the three above criteria) of cardinality constraints to CNF  based on boolean adders using the unary representation of integers. 

Many developments have been done since : refinements of the translation of cardinality constraints, CNF translation of general arithmetic constraints... It would be difficult to cite them without omitting some. Some papers studied the existence of good translations to CNF as a target Constraint. That is the subject we choose to tackle and generalize in this paper through some incomplete thoughts.

A breakthrough was made in \cite{Bessiere2009CircuitCA} by giving a necessary and sufficient condition for a source Constraint to have a good translation to CNF. The condition is that a propagator that enforces generalized arc consistency on this source Constraint must be computed by a polynomial size monotone circuit. As a consequence of this result, a superpolynomial lower bound for the monotone circuit that encodes a propagator for some Constraint implies that there is no good translation to CNF for this Constraint. It is the case of ALLDIFF as mentioned \cite{Bessiere2009CircuitCA} and also for its generalization GCC.

In the same vein, \cite{GMKO2014} proves that there exists no GAC translation of systems of an arbitrary number of XOR equations.  By contrast, it gives a good translation to CNF for XOR systems having a constant number of equations. 

These results draw the outlines of classes of Constraints according to the existence of good translations to CNF. We describe in the following a classification that does not consider necessarily CNF as a target Constraint and we discuss it.

\section{Comparing constraints according to their propagation power \label{sec:classification}}

Polynomial time constraints translation, in its simplest acceptance, i.e., the target constraint network and the source constraint network must be equiconsistent, is a Karp-reduction in the sense of complexity theory.
It allows to polynomialy translate a source problem into an equivalent target problem. 
We propose to call \textit{GAC-reduction} any Constraint reduction that preserves the deductive power of domain consistency enforcement. Such a reduction allows to categorize Constraints with respect to their propagation powers.

Let us illustrate the notions of reduction and GAC-reduction with an example involving two Constrains : CNF, namely the propositional clauses, and $\neq$, namely the binary difference constraints on finite domains.

\begin{figure}[ht]
\centering
\centerline{\includegraphics[width=0.5\textwidth]{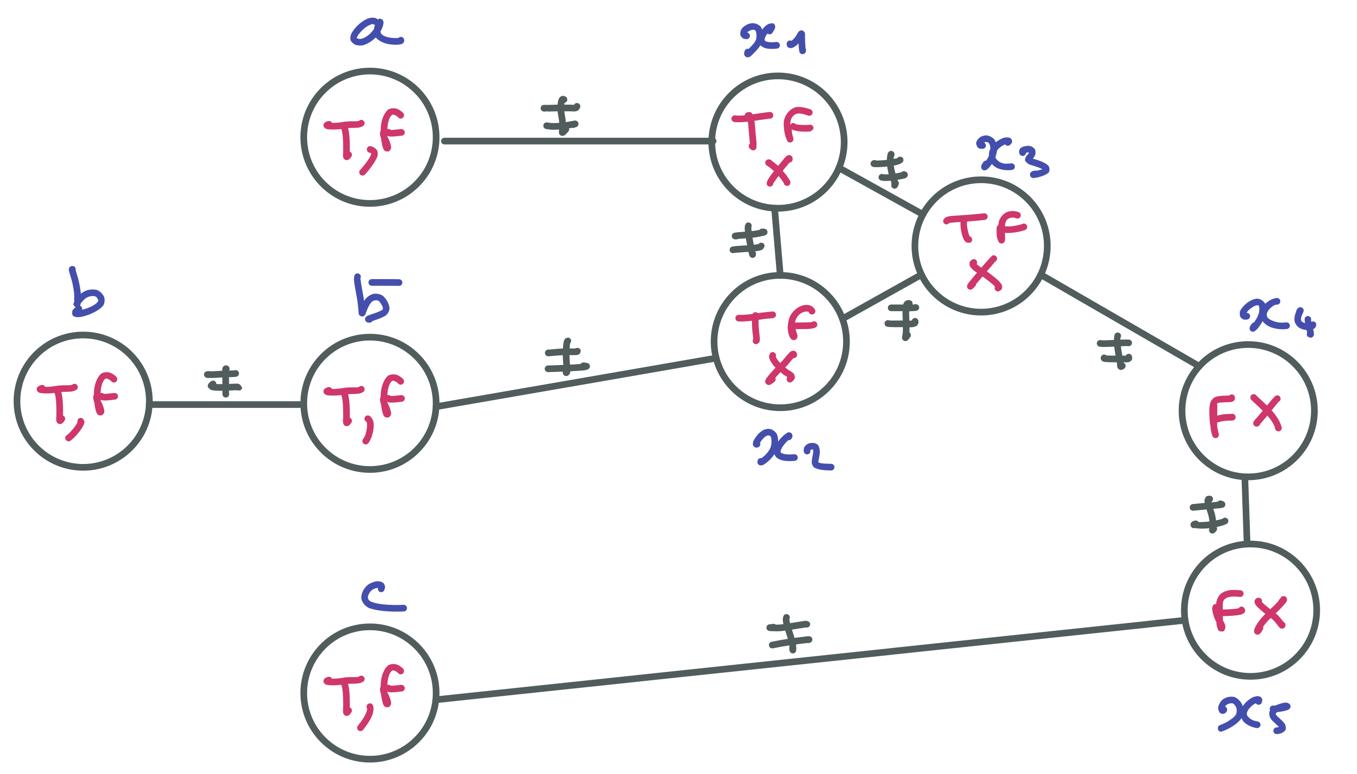}}
\caption{A (non-GAC) reduction from the clause $(a \vee \neg b \vee c)$ to binary difference constraints.}
\label{fig:cnf2dif}
\end{figure}

Figure \ref{fig:cnf2dif} shows a reduction from CNF to $\neq$ in the case where the source constraint is a clause of 3 literals, but which can be generalized to a clause of any length.

The variables $a, b, c$ of the resulting target constraint network correspond to the variables of the source clause $(a \vee \neg b \vee c)$, with domains $\{\text{F},\text{T}\}$, where T stand for true and F for false. The variables $x_1$ to $x_5$ are called auxiliary variables. If $a=\text{F}$, $b=\text{T}$ and $c=\text{F}$ the target network cannot be satisfied. For all other assignations of $a, b, c$, it can be satisfied.

But this reduction is \textit{not} a GAC-reduction. For example, with $c=\text{F}$, $b=\text{T}$ and $a$ unassigned, unit propagation, which enforces domain consistency on the source clause, assigns $a=\text{T}$. But on the target network, enforcing domain consistency on the difference constraints does \textit{not} assign $a$.

\begin{figure}[ht]
\centering
\centerline{\includegraphics[width=0.65\textwidth]{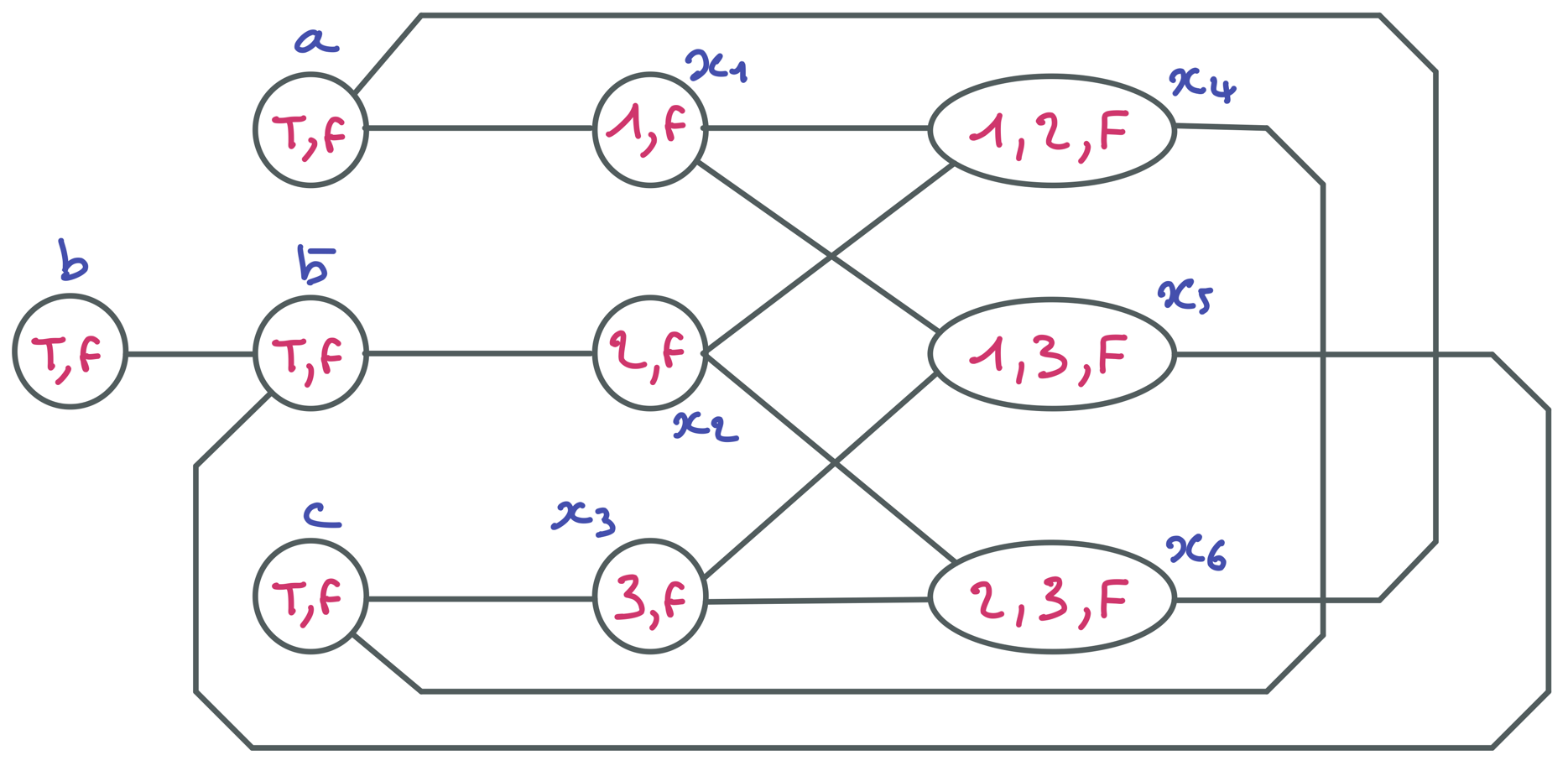}}
\caption{A GAC-reduction from the clause $(a \vee \neg b \vee c)$ to binary difference constraints.}
\label{fig:GACcnf2dif}
\end{figure}

On the other hand, the reduction from CNF to $\neq$ presented figure in \ref{fig:GACcnf2dif} on the same example \textit{is} a GAC-reduction. If the source clause is falsified, then the target network in inconsistent and enforcing domain consistency on its constraints produces an empty domain. In addition to that, in the three cases where unit propagation on the source clause assigns a variable, enforcing domain consistency on  constraints of the target network do the same assignation.

Incidentally, if we want to solve SAT instances with a graph coloring solver, it might be more efficient to use the second reduction rather than the first one.

GAC-reduction may involve different domains for the source variables and the related target variables. For example, to reduce a constraint $A \neq B$ in CNF, where $A$ and $B$ have domains $1..n$, the source variable A (B) can be represented in the target network by $n$ variables $a_1,...,a_n$ ($b_1,...,b_n$) with domains $\{\text{T},\text{F}\}$, which is expressive enough to represent not only all possible assignments of A and B, but also all possible restrictions of their domains.

The GAC-reduction can be completed as follow:
\begin{enumerate}
\item A set card[1..1]($a_1,...,a_n$) of clauses encoding that exactly one variable among $a_1,...,a_n$ must be set to 1, see for example \cite{Sinz2005TowardsAO} \cite{Hlldobler2013AnEE} for different GAC-compliant solutions.
\item A set card[1..1]($b_1,...,b_n$) of clauses encoding that exactly one variable among $b_1,...,b_n$ must be set to 1.
\item The clauses $(\neg a_1 \vee \neg b_1), ..., (\neg a_n \vee \neg b_n)$.
\end{enumerate}

Here, not only GAC-reduction can introduce auxiliary variables (depending of the encoding of the card[1..1] constraints), but it may also use different representations of the possible restrictions of source domains and the corresponding target domains.

\begin{figure}[ht]
\centering
\centerline{\includegraphics[width=0.7\textwidth]{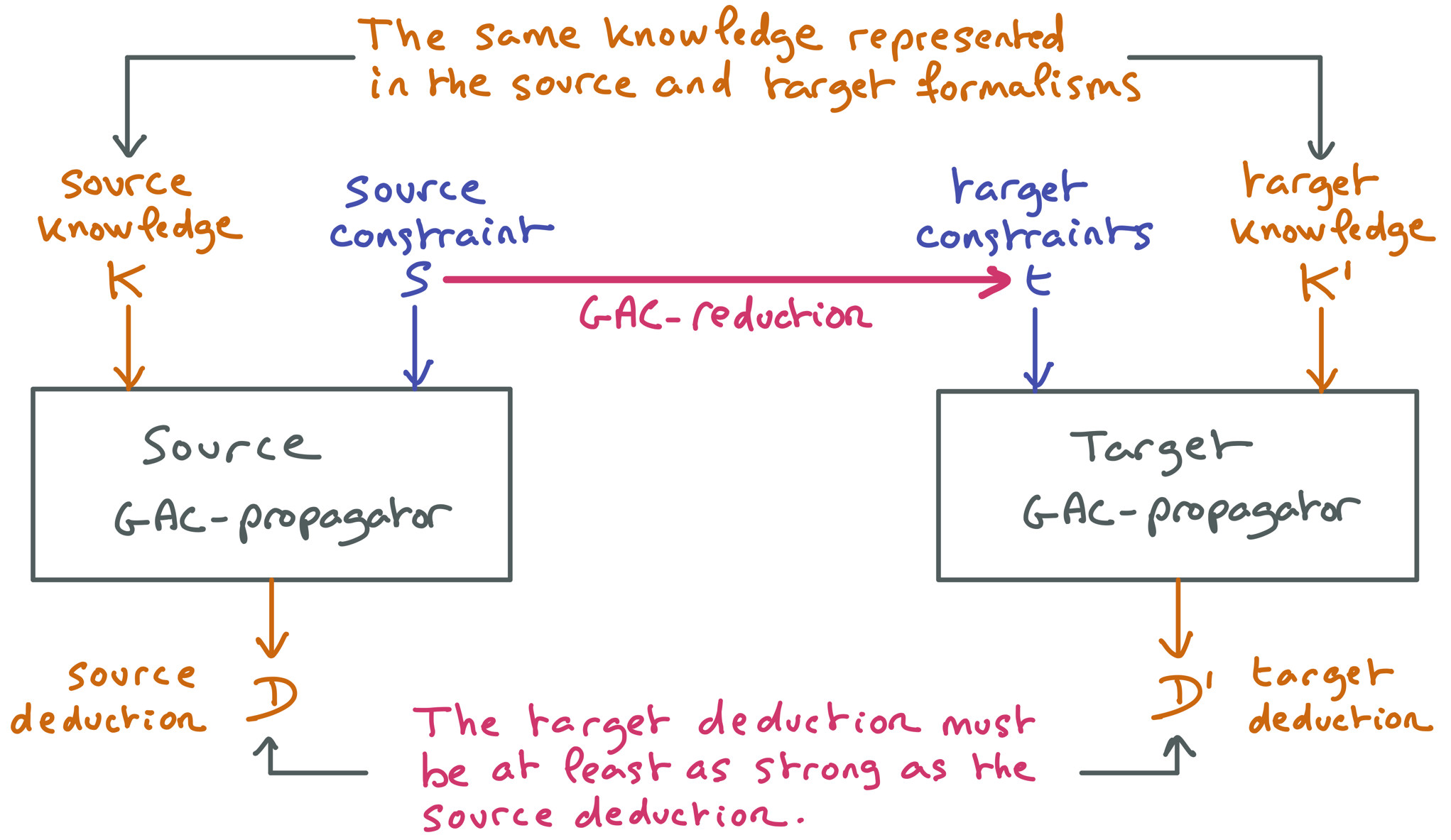}}
\caption{A schematic representation  a GAC-reduction.}
\label{fig:GACreduction}
\end{figure}

A formal definition of a GAC-reduction - which we will not give here - should express the idea, summarized in figure \ref{fig:GACreduction} that if, from a certain source knowledge K, the GAC-propagation of the source constraint produces a deduction D, then from the same knowledge expressed in the target representation, the GAC-propagation of the target constraint network should produce a deduction at least as strong as D.
K and D are restrictions of the domains of the source variables. The corresponding target knowledge and deduction are restrictions of domains of target variables. A domain $d$ is a restriction of another domain $k$ if $d \subseteq k$.

The main result in \cite{Bessiere2009CircuitCA} can be reformulated in terms of GAC-reduction this way: for any finite domains Constraint $S$, the GAC-propagator of $S$ can be implemented by a monotone Boolean circuit if and only if there is a GAC-reduction from $S$ to CNF.
We identify here a class of Constraints that we call MPC (for Monotone Polynomially Propagatable Constraints), which is the set of all the Constraints that can be GAC-reduced to CNF. De facto, CNF is a MPC-complete Constraint.

All the generalizations of CNF belonging to MPC, such as Boolean cardinality Constraint (Card) and pseudo-Boolean inequalities (PB$_>$), are obviously MPC-complete. (Card $\in$ MPC follows from the GAC-reduction proposed in \cite{BB03} and PB$_>$ $\in$ MPC follows from the one proposed in \cite{BBR09}). 

It can be noticed that a necessary condition for a type Constraint $Q$ be MPC-complete is that $Q$ is NP-complete - in the sense that the problem of determining the consistency of a constraint network of type $Q$ is NP-complete - because the existence of a GAC-reduction from $Q$ to CNF implies the existence of a Karp-reduction from $Q$ to CNF. 
So the XOR Constraint, which belongs to MPC but is not NP-complete, is therefore not MPC-complete.

Also according to \cite{Bessiere2009CircuitCA}, there are Constraints, such as ALLDIFF and its generalization GCC (Global Cardinality Constraint \cite{GCC}), whose CAG-propagators can be computed in polynomial time, but which are not in MPC. We propose to call PPC (for Polynomially Propagatable Constraints) the set of all finite domain Constraints whose GAC-propagators have polynomial time complexity, and then can be implemented as polynomialy sized Boolean circuits. De facto, MPC is strictly included in PPC, and ALLDIFF and GCC are in PPC $\backslash$ MPC. We do not know if there are PPC-complete Constraints, and the question seems interesting to us insofar as a solver optimized for such a Constraint could perhaps have a broader application scope than any solver dedicated to a MPC Constraint, such as SAT solvers or even pseudo-Boolean solvers.

Finally, it should be noted that there are Constraints whose satisfaction is verified in polynomial time, but for which enforcing domain consistency is a NP-hard problem. This holds with pseudo-Boolean inequalities (PB$_=$), for example. Under the assumption P $\neq$ NP, these Constraints are not in PPC. We propose to call PVC the set of polynomialy verifiable Constraints, in the sense that a finite domain Constraint $Q$ is in VPC if and only if there is a polynomial time complexity algorithm that checks whether a complete assignment satisfies a constraint $q$ of type $Q$.

\begin{figure}[ht]
\centering
\centerline{\includegraphics[width=0.7\textwidth]{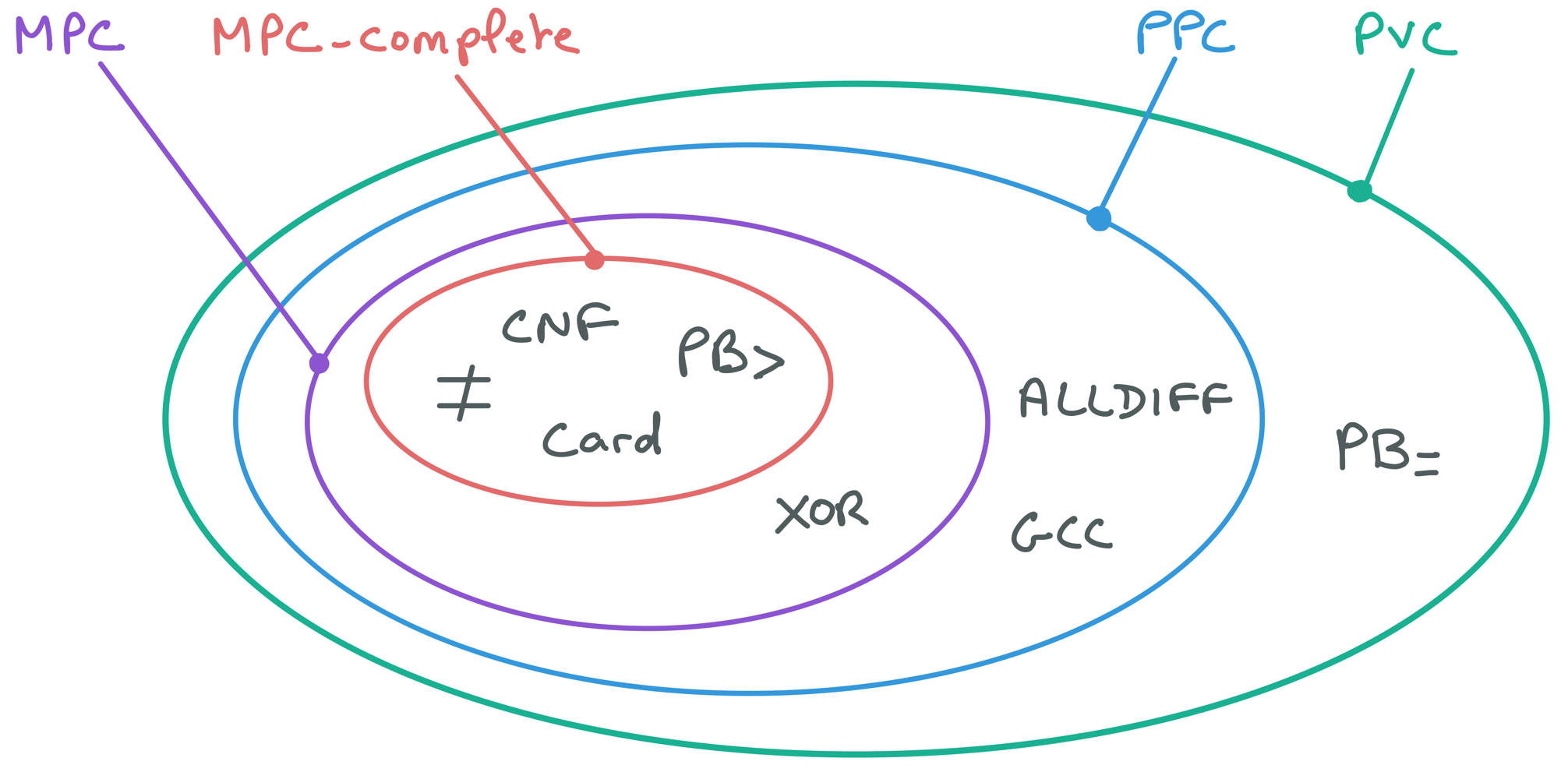}}
\caption{A classification of \textsc{C}onstraints}
\label{fig:classes}
\end{figure}

Figure \ref{fig:classes} shows the classes MPC, PPC and VPC, as well as some Constraints belonging to these classes.

\section{Concluding remarks and questions}

In a world where solvers are based on enumeration and domain consistency enforcement, a solver dedicated to a Constraint of the PPC $\backslash$ MPC class has a higher deductive power than a solver dedicated to an MPC Constraint, such as a SAT solver, a graph coloring solver, or even a pseudo-Boolean solver. And in such a world, CNF is a poor candidate for the role of universal Constraint into which all polynomially propagatable Constraints could be translated without loss of resolution efficiency. If we were to look for a universal solver, we would first have to identify the complete constraints of the PPC class. In the current state of our young reflections, we do not even know if such Constraints exist and so one question that could be addressed would be: are there Constraints complete for the class PPC?

But in a world where solvers use more sophisticated deductive systems, things are becoming more complex. In \cite{GwynneK14} is addressed the question of power of deduction restricted to the representations of boolean functions but they consider also more  sophisticated systems of deduction than GAC. The same can be done with Constraints with general domains. For example, the GAC-propagation of Constraints beyond MPC, such as GCC, cannot be polynomially simulated by unit propagation in a SAT solver or by GAC-propagation in a pseudo-Boolean solver. But what if the SAT solver replaces unit propagation with bounded resolution or if the pseudo-Boolean solver uses cutting planes?

Finally, we think that the study of good practices of constraint translation will have to be strongly connected to the study of solvers in order to progress.

\bibliography{main}
\bibliographystyle{alpha}
\newpage

\end{document}